\documentclass{article}

\usepackage{multicol}
\usepackage{authblk}
\usepackage{geometry}
\usepackage{graphicx}
\usepackage{url}
\usepackage{subcaption}
\usepackage{float}

\title{Inpainting the Red Planet: Diffusion Models for the Reconstruction of Martian Environments in Virtual Reality}
\author[1]{Giuseppe Lorenzo Catalano}
\author[1]{Agata Marta Soccini}
\affil[1]{Computer Science Department, Universit\`a degli Studi di Torino, Corso Svizzera 185, Torino, 10149, Italy}
\date{}

\begin{document}
\maketitle

\begin{figure*}[h!]
    \centering
    \includegraphics[width=0.8\linewidth]{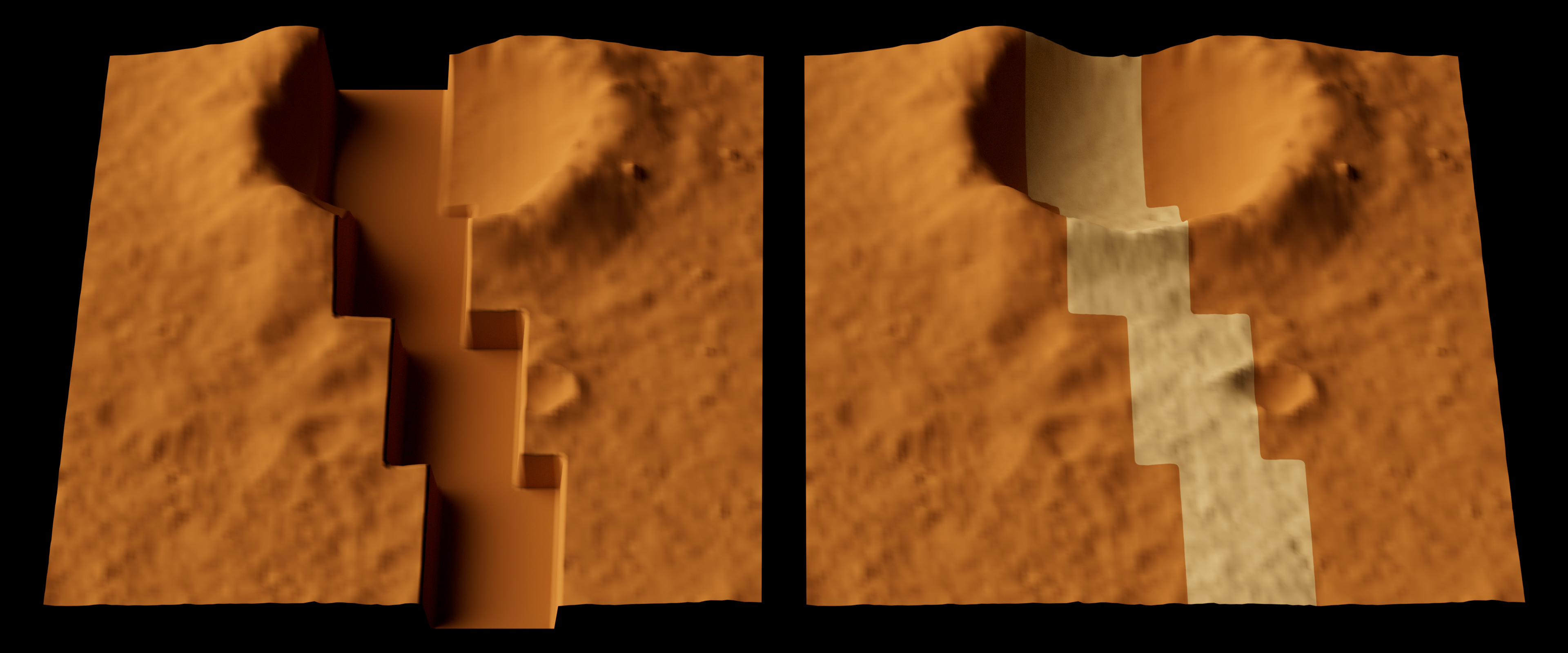}
    \caption{The generative capabilities of diffusion models provide realistic restorations of degraded acquisitions, producing structurally consistent representations suitable for virtual environments. (Left) A degraded Martian heightmap depicting impact craters, with regions composed of missing values. (Right) The same heightmap after reconstruction, using our diffusion-based approach; the inpainted regions are explicitly highlighted.}
    \label{fig:teaser}
\end{figure*}

\begin{multicols}{2}

\begin{abstract}
Space exploration increasingly relies on Virtual Reality for several tasks, such as mission planning, multidisciplinary scientific analysis, and astronaut training. A key factor for the reliability of the simulations is having accurate 3D representations of planetary terrains. Extraterrestrial heightmaps derived from satellite imagery often contain missing values due to acquisition and transmission constraints.
Mars is among the most studied planets beyond Earth, and its extensive terrain datasets make the Martian surface reconstruction a valuable task, although many areas remain unmapped.
Deep learning algorithms can support void-filling tasks; however, whereas Earth's comprehensive datasets enables the use of conditional methods, such approaches cannot be applied to Mars. Current approaches rely on simpler interpolation techniques which, however, often fail to preserve geometric coherence.
In this work, we propose a method for reconstructing the surface of Mars based on an unconditional diffusion model. Training was conducted on an augmented dataset of 12000 Martian heightmaps derived from NASA's HiRISE survey. A non-homogeneous rescaling strategy captures terrain features across multiple scales before resizing to a fixed $128 \times 128$ model resolution. We compared our method against established void-filling and inpainting techniques, including Inverse Distance Weighting, kriging, and Navier-Stokes algorithm, on an evaluation set of 1000 samples. Results show that our approach consistently outperforms these methods in terms of reconstruction accuracy (4-15\% on RMSE) and perceptual similarity (29-81\% on LPIPS) with the original data.
\end{abstract}

\section{Introduction}

The benefits of Virtual Reality (VR) for space exploration are being investigated with growing interest in the recent years \cite{Holt2023}. Use cases include the training of future astronauts \cite{Garcia2020}, the remote operation of existing rovers \cite{CastillaArquillo2024} and the interactive real-time visualization of spacecraft diagnostics in immersive environments \cite{Soccini1, Soccini2}. Additionally, such technologies can enable detailed morphological studies of extraterrestrial surfaces for mission planning \cite{Garcia2019}. Planet Mars specifically is an active object of study by the scientific community, and realistic representations of Martian landscapes is vital for the success of these missions.

To be effective, many of these applications rely on the ability to integrate and accurately represent the terrains of various celestial bodies in VR \cite{Catalano2024}. Surfaces are commonly represented as heightmaps, a type of raster grid in which each pixel encodes a specific altitude value, that are particularly well suited for representing structured data in immersive virtual environments \cite{Kraus2021}.

The problem we address is that most of the data sources regarding extraterrestrial surfaces are captured by spacecraft, such as the Mars Reconnaissance Orbiter \cite{Zurek2007}, thus presenting additional challenges in being acquired and transmitted given the difficult environment faced by the instruments. These factors contribute to a scenario in which, unlike Earth where detailed and abundant measurements are often accessible, significantly less information is available. As a consequence, the final processed heightmaps may present missing values within the raster. This can be the consequence of errors during the acquisition of the original signal, data loss during transmission from the orbiter, or ambiguities arising during the processing phase that converts images into heightmaps. In such cases, performing another full acquisition of the raw data may be prohibitively expensive, or even impossible if the dataset refers to past events. 

Data gaps can severely hinder the usability of heightmaps for research and mission planning purposes. To face the problem, we can use void-filling methods in order to restore the invalid regions through suitable approximations. These methods typically rely on interpolation techniques or statistical analyses of the valid neighbourhoods to estimate the missing values \cite{Crema2020}. Moreover, given the raster-based structure of heightmaps, which can be encoded and visualized as images, a wide range of inpainting algorithms for the modification or reconstruction of selected regions within a digital image can be leveraged as well \cite{Bertalmio2000}.
Recently, diffusion models \cite{Sohl2015, Ho2020} have outperformed other deep learning-based approaches for image generation, such as Generative Adversarial Networks (GANs) \cite{Goodfellow2020} and Variational Autoencoders (VAEs) \cite{Kingma2013}. These models generate content by reversing a Gaussian diffusion process that progressively adds noise to the input data. Diffusion models have shown superior performance in generative tasks \cite{Dhariwal2021}, and have rapidly become the state of the art in many image-related applications; nonetheless, active research is being conducted in order to assess the performance of such architectures in relation to other kinds of data \cite{Chen2024}.

Diffusion models have already been studied for heightmap generation and restoration as well \cite{Lo2024, Zhao2024}. However, these approaches are not suitable for the reconstruction of Mars because: i) these models were trained and tested only on terrains related to limited regions of Earth, making them potentially susceptible to overfitting over specific geometrical features, and not necessarily suitable for other celestial bodies; ii) these approaches rely on conditional generation, supposedly bringing further overfitting on specific data distributions. Whereas this approach may be effective for Earth-related tasks, such auxiliary information is often unavailable in the case of Mars; iii) the suggested approaches do not evaluate their results from a visual standpoint as well, only resorting on quantitative assessments and brief overviews in 2D that say little in terms of the reconstruction quality for ultimately using them in VR.

In this context, we applied our method based on unconditional diffusion models, namely models that do not require further contextual information during generation, for heightmap reconstruction; our hypotheses were formulated as follows:

\begin{itemize}
    \item \textbf{H1.} Our method restores Martian surfaces smoothly and consistently, producing visually coherent geometrical features that are well-suited for 3D visualization and simulations in virtual reality.
    \item \textbf{H2.} The digitally reconstructed Mars surfaces are objectively similar to the original terrains, even though data sources are particularly scarce.
\end{itemize}

The contributions of this work can be summarized as follows.
I) Unconditional diffusion-based reconstruction framework for planetary terrain inpainting, addressing the lack of auxiliary conditioning data for extraterrestrial surfaces such as Mars.
II) Augmented Martian terrain dataset derived from NASA's HiRISE survey, with training which incorporates a non-homogeneous rescaling strategy to capture multi-scale terrain features before standardizing resolution to $128 \times 128$.
III) Comprehensive evaluation against void-filling and inpainting methods (Inverse Distance Weighting, kriging, and Navier-Stokes), using both quantitative error metrics (RMSE, MAE, PSNR, EMD) and perceptual metrics (LPIPS, SSIM, FID).
IV) Performance gains of up to $15\%$ in RMSE and $81\%$ in LPIPS, producing geometrically consistent reconstructions suitable for VR-based planetary visualization and analysis.

The paper is organized as follows: Section \ref{sec:related} reviews related work, Section \ref{sec:methods} presents the proposed method, Section \ref{sec:experiments} describes the experiments, Section \ref{sec:results} reports the results, Section \ref{sec:discussion} discusses the findings, and Section \ref{sec:conclusions} concludes the paper.

\section{Related Work}
\label{sec:related}

\subsection{Diffusion Models}

Diffusion models \cite{Sohl2015} are a family of generative methods that are based on reversing a diffusion process, namely a procedure that progressively adds noise to a signal over a series of discrete timesteps, until the original input is transformed into pure noise. By learning the reverse of this process, it becomes possible to generate new synthetic data starting from random noise. The forward diffusion is modeled as a discrete-time Markov chain that adds Gaussian noise across timesteps $t = 0, \; \dots , \; T$ where $t = 0$ corresponds to an image without noise and $t = T$ is pure Gaussian noise. Each intermediate corrupted sample $x_t$ can be expressed in closed form, with regards to the original image $x_0$ and the timestep $t$: $x_t \sim \mathcal{N}(\sqrt{\bar{\alpha}_t} x_0), (1 - \bar{\alpha}_t) \mathbf{I})$, where $\bar{\alpha_t}$ is a set of fixed constants. This formulation allows for efficient training, as corrupted samples can be directly generated from clean data without simulating the entire chain.
In recent years, several mathematical formulations and frameworks based on this principle have emerged, particularly in the domain of image synthesis. One of the most prominent examples is given by Denoising Diffusion Probabilistic Models (DDPMs) \cite{Ho2020}, in which the actual denoising model, typically a U-Net \cite{Ronneberger2015}, is trained to predict the noise component at each timestep. The generation process begins by sampling $x_T$ from a random distribution, $x_T \sim \mathcal{N}(\mathbf{0}, \mathbf{I})$, and iteratively removing the predicted noise through the learned reverse process. The procedure ends when $x_0$, a synthetic image without noise resembling the distribution of the training data, is produced.

Subsequent works focus on speeding up the inference process, which requires the entire diffusion chain to be traversed in the case of DDPMs. An example is given by Denoising Diffusion Implicit Models (DDIMs) \cite{Song2020}, which reformulate the mathematical process but keep the same training objective, making it possible to skip inference of some timesteps while keeping the same trained denoising networks. Another example is given by Latent Diffusion Models (LDMs) \cite{Rombach2022}, which leverage a pre-trained autoencoder to perform denoising in the latent space. With these regards, Diffusion Transformers (DiTs) \cite{Peebles2023} build upon previous works by leveraging the Transformer \cite{Vaswani2017} architecture as the noise predictor in the DDPM formulation.

\subsection{Surface Void-Filling Techniques}
\label{sec:related_voidfilling}

To restore invalid regions within a heightmap, various approaches can be employed to interpolate the missing values between valid pixels. The general idea is to exploit the spatial information contained in neighboring values to estimate the unknown ones \cite{Crema2020}. Simpler methods are based on nearest-neighbor strategies or spline interpolation, while more advanced techniques involve weighted estimations. Among the latter, Inverse Distance Weighting (IDW) and kriging are two of the most commonly used.
IDW \cite{Shepard1968} computes a weighted average of nearby valid pixels, assigning greater influence to those closer to the missing point. Kriging \cite{McBratney1986} produces a weighted average as well, but the weights are derived from a statistical model of spatial autocorrelation through the use of a semi-variogram.
The heightmaps that are part of the HiRISE dataset, the main source for Martian terrains (which is further discussed in Section \ref{sec:methods_dataset}), are processed by linearly interpolating the areas with too much value uncertainty \cite{Sutton2022}.
In general, the performance of each method depends on the morphological characteristics of the surface patch that is being reconstructed \cite{Reuter2007}. 

Deep learning generative models have been successfully employed for surface restoration. GANs were tested in recent years in the context of terrain surfaces \cite{Gavriil2019, Yan2021, Zhou2022}; diffusion models have also started to be analyzed in this context as well \cite{Lo2024, Zhao2024}. As previously mentioned, one major drawback of such methods is that they rely on a conditioned approach, where the generation process is guided by supplementary data; more specifically, conditioning is performed using the shape of the binary mask \cite{Lo2024}, or even the general structure of the terrain that must be known beforehand \cite{Zhao2024}. Moreover, the studies were performed on heightmaps representing specific areas of Earth. These reasons may lead to overfitting over small use cases, also limiting the generalization of the approaches to different scenarios, making their use for Mars more challenging. Finally, the studies were not concerned about the visual inspection of the restoration output, with a focus that strayed from the representation in VR. Our aim with these regards is to define a framework that can be applied in diverse use cases, applicable to terrain morphology but also for other purposes, which do not require sets of additional data of different kind.

\subsection{Image Inpainting Algorithms}
\label{sec:related_inpainting}

Inpainting is a well-known task in the digital image processing domain. The goal is to fill specific portions of an image, typically selected using a binary mask, with new values that are coherent with the adjacent valid portions. This task is usually performed to remove unwanted features or to restore partial degradations.
Several inpainting algorithms have been developed throughout the years, based on different principles, such as the fast marching method \cite{Telea2004} or the Navier-Stokes fluido-dynamics equations \cite{Bertalmio2001}, to expand the valid parts of the signal into the missing regions. Other methods rely instead on computing patch correspondences for finding the best match \cite{Barnes2009}.
The advent of generative deep learning models for image manipulation had a consistent impact for the task of inpainting as well \cite{Quan2024}. Diffusion model-based methods were successfully implemented for this purpose as well. Notable example include Palette \cite{Saharia2022} and Stable Diffusion \cite{Rombach2022}, however both rely on a generation that is conditioned by the inpainting binary mask.
A different approach is presented by RePaint \cite{Lugmayr2022}, an inpainting algorithm which is based on unconditioned diffusion models. More specifically, a pre-trained DDPM is used for performing the unconditional reverse diffusion process; by itself, this process would produce a result with no correlation with the image that needs to be inpainted.  After each denoising step, however, the valid parts of the image are overwritten on the intermediate result, so that the next step will be guided towards generating a final result that is coherent with the original input. In order to obtain better results, a resampling mechanism is also implemented; noise is added and then removed multiple times, in order to generate patches that are more consistent with the valid parts. Given its flexibility with regard to the inpainting masks, guaranteed by the unconditional DDPM, we based our method on this algorithm.

To the best of our knowledge, this is the first work to address the specific task of Martian terrain restoration using diffusion models, achieving performance superior to currently employed methods. Our approach leverages the absence of additional data constraints afforded by unconditional models to tackle the unique challenges of Martian datasets. Furthermore, the model is trained on features sampled at multiple scales from across the entire planet, enhancing its ability to generate accurate and coherent terrain features.

\section{Methods}
\label{sec:methods}

\subsection{Martian Surfaces Dataset}
\label{sec:methods_dataset}

The training of the DDPM model required a properly structured dataset of Martian terrain heightmaps. The most prominent repository with this regard is provided by HiRISE \cite{McEwen2007}, a set of high-performance cameras mounted on the Mars Reconnaissance Orbiter (MRO), dedicated to capturing high-resolution imagery of the planet’s surface. HiRISE is capable of acquiring stereoscopic image pairs, which undergo a dedicated processing pipeline to extract the corresponding heightmaps \cite{Kirk2008, Sutton2022}.
The resulting data products are made publicly available via an online repository\footnote{\url{https://www.uahirise.org/dtm/}} in the form of Digital Elevation Models (DEMs), a specific type of terrain representation that includes metadata for geo-localization, such as coordinate reference systems and coordinates of each acquisition.

\begin{figure}[H]
    \centering
    \begin{subfigure}{0.4246\linewidth}
        \includegraphics[width=\textwidth]{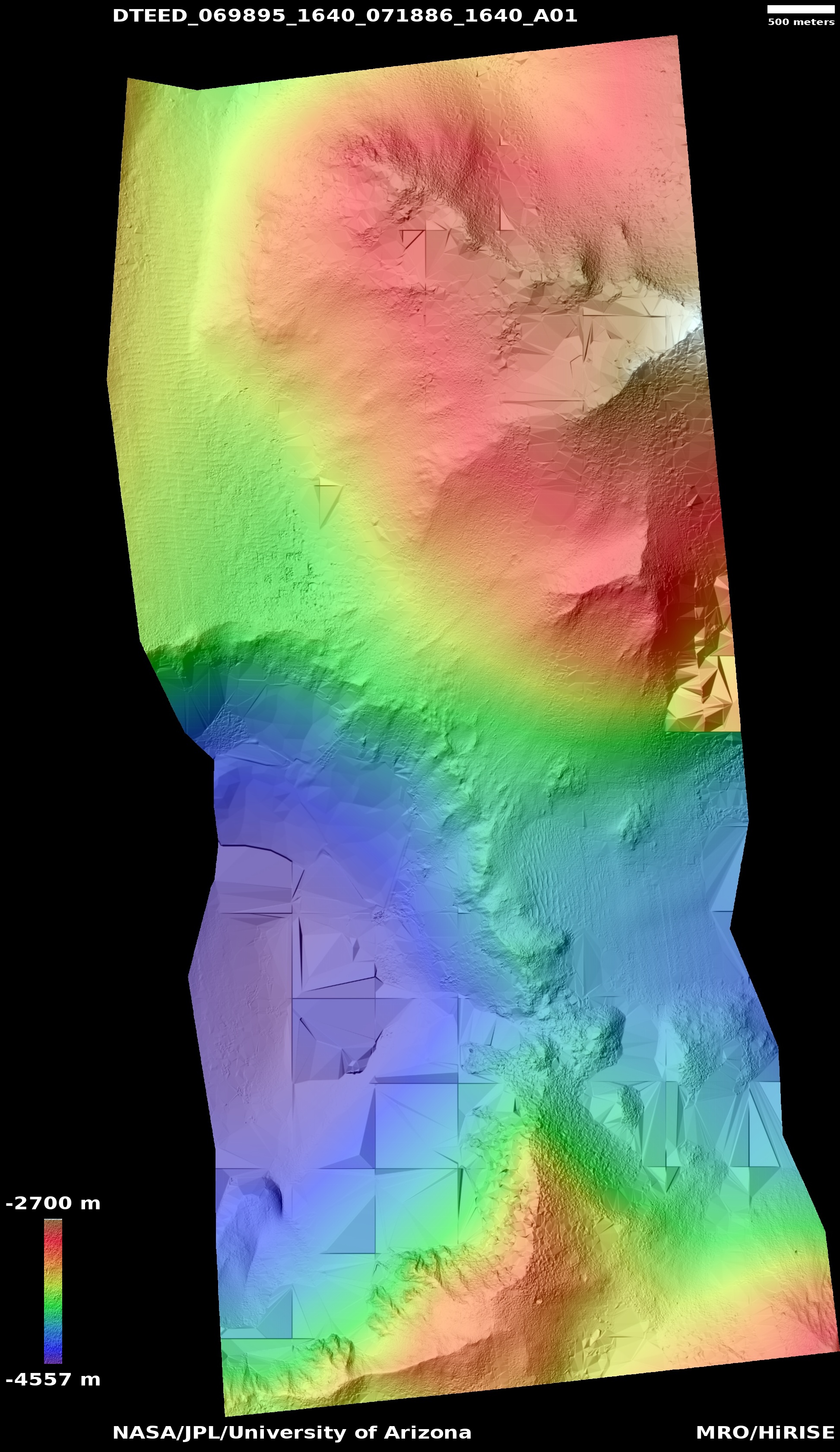}
        \label{fig:hirise_1}
    \end{subfigure}
    \begin{subfigure}{0.2\linewidth}
    \end{subfigure}
    \begin{subfigure}{0.4\linewidth}
        \includegraphics[width=\textwidth]{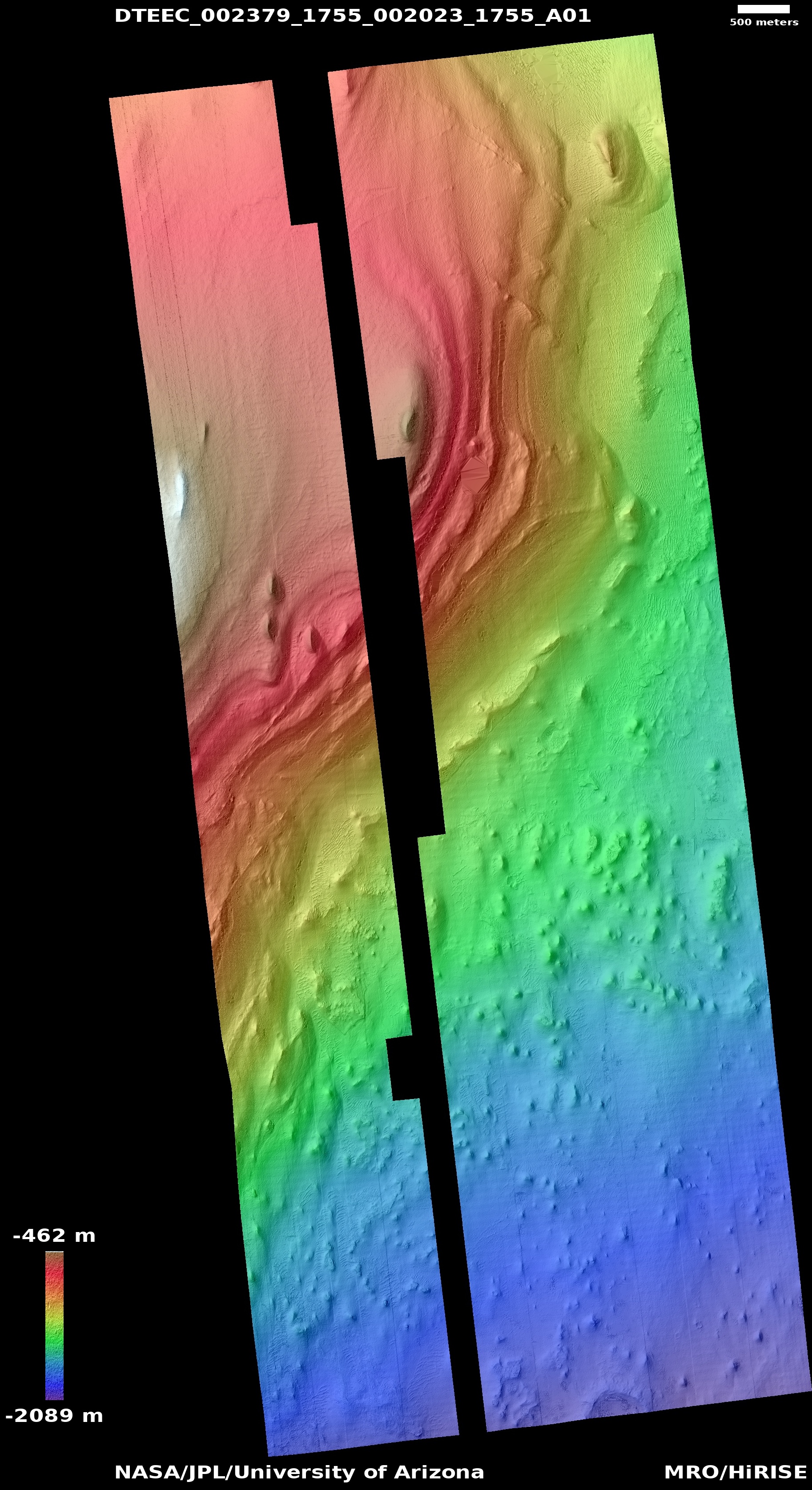}
        \label{fig:hirise_2}
    \end{subfigure}
    \hfill
    \caption{RGB representation of Martian surfaces, obtained from the processing of HiRISE stereo pairs. It may occur that missing data are interpolated during processing (the squared patterns within the crater in the left image) or entirely unavailable (the black line in the right image).}
    \label{fig:hirise}
\end{figure}

These terrains are a valuable source for investigating the morphology of the planet, given their capability to represent detailed patterns due to their high resolution \cite{Goudge2017, QuantinNataf2021}.
This use case presents an interesting testbed: due to the challenges faced during data acquisition and transmission, certain regions of the images may be difficult to process or entirely unavailable, thus resulting in interpolated or missing values as already discussed (see Figure \ref{fig:hirise}).
Approximately 1,150 heightmaps are available at the current time. However, these are not immediately suitable for direct use in training. Indeed, diffusion models require normalized images of fixed size ($128 \times 128$ in our case) as input, whereas HiRISE DEMs typically cover much larger areas, in the order of $10^3$ pixels per side. Furthermore, the data are stored in the Planetary Data System (PDS) data format, which is not supported by most image processing libraries.
For these reasons, a new dataset was generated to enhance file interoperability and readability. The resulting training set consists of 12,000 normalized heightmaps, stored using the TIFF file format. Each element is a random crop extracted from the original HiRISE survey, with side lengths randomly selected between 512 and 2048 pixels for each sample. During training, these crops are rescaled in a non-homogeneous way to $128 \times 128$, using nearest-neighbour interpolation. The main reason behind this choice, made mainly to support the investigation of \textbf{H2}, is to provide the model with knowledge of terrain features that may develop over portions that are larger than the model resolution. This operation enables data augmentation, helping to compensate for the relatively small size of the initial dataset which is an an inherent challenge in this domain. GDAL\footnote{\url{https://gdal.org/en/stable/}} was used to process the original PDS data and convert them to TIFF.

\subsection{Training of the Diffusion Model}
\label{sec:methods_training}

In order to assess both \textbf{H1} and \textbf{H2}, we trained an unconditional DDPM model, capable of generating realistic Martian terrains without the need for textual or visual prompts. To this purpose, training  was performed at a resolution of $128 \times 128$ pixels, using the augmented dataset described in Section \ref{sec:methods_dataset}. As previously mentioned, we believe that by feeding the model random crops of varying original size, the model can learn surface features at multiple different scales.

\begin{figure}[H]
    \centering
    \includegraphics[width=\linewidth]{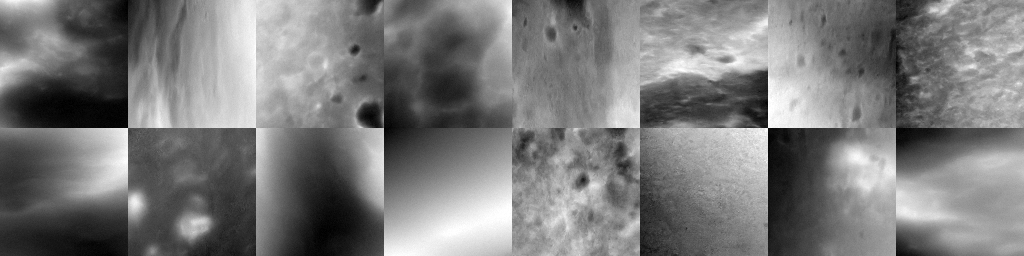}
    \caption{Evaluation of the trained unconditional DDPM after 100 epochs of training with Martian terrains.}
    \label{fig:evaluation}
\end{figure}

Training was conducted over 100 epochs, with a batch size of 16 samples and a total number of $T = 1000$ timesteps. The main neural network specifications, such as the model architecture and loss function, were chosen according to the original DDPM formulation \cite{Ho2020}; specifically, we employed a U-Net to predict the total noise in an image at an arbitrary timestep of the diffusion process. The total training time was approximately 8 hours, using 2 NVIDIA A40 GPUs. Samples from the trained DDPM after 100 epochs are shown in Figure \ref{fig:evaluation}. The implementation used for DDPMs is based on the Huggingface Diffusers\footnote{\url{https://huggingface.co/docs/diffusers}} library.

\subsection{Inference}
\label{sec:methods_inference}

\begin{figure}[H]
    \centering
    \includegraphics[width=0.75\linewidth]{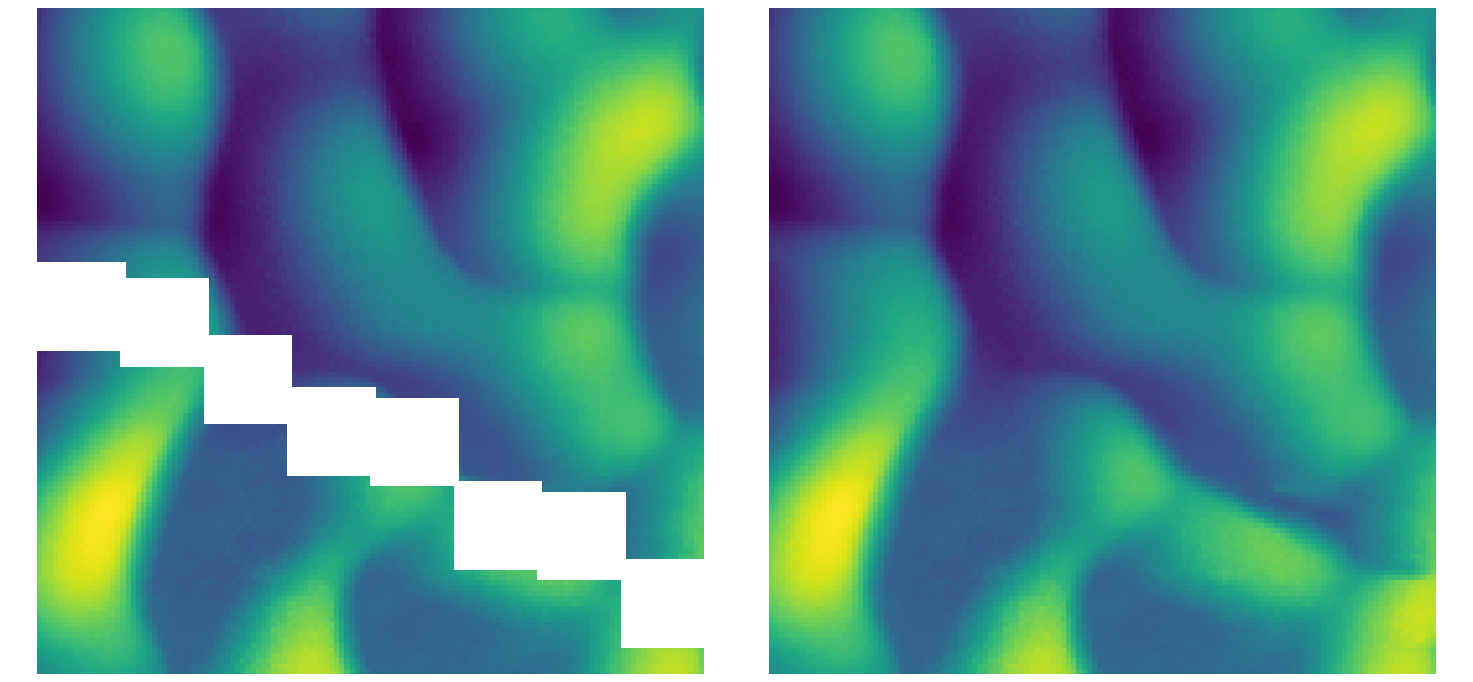}
    \caption{Input (left) and output (right) of our approach to Martian surface reconstruction, represented as 2D rasters.}
    \label{fig:plt_in_out}
\end{figure}

Inference with the trained DDPM model is performed by using the RePaint algorithm on normalized heightmaps. Specifically, a selected portion of Martian terrain is resized to the fixed resolution of $128 \times 128$ pixels and given as input, along with a binary mask indicating which parts need to be restored. The main parameter configuration suggested by the original authors was employed; the complete inference takes around 44 seconds on a single NVIDIA A40 GPU.
It is important to highlight the role of the mask in the inference process: while it is indeed used by the inpainting algorithm to distinguish between valid and missing pixels, it is however not directly used by the diffusion model itself, which operates unconditionally. Figure \ref{fig:plt_in_out} illustrates the input and output of our method.

\section{Experiments}
\label{sec:experiments}

\begin{figure}[H]
    \centering
    \includegraphics[width=0.8\linewidth]{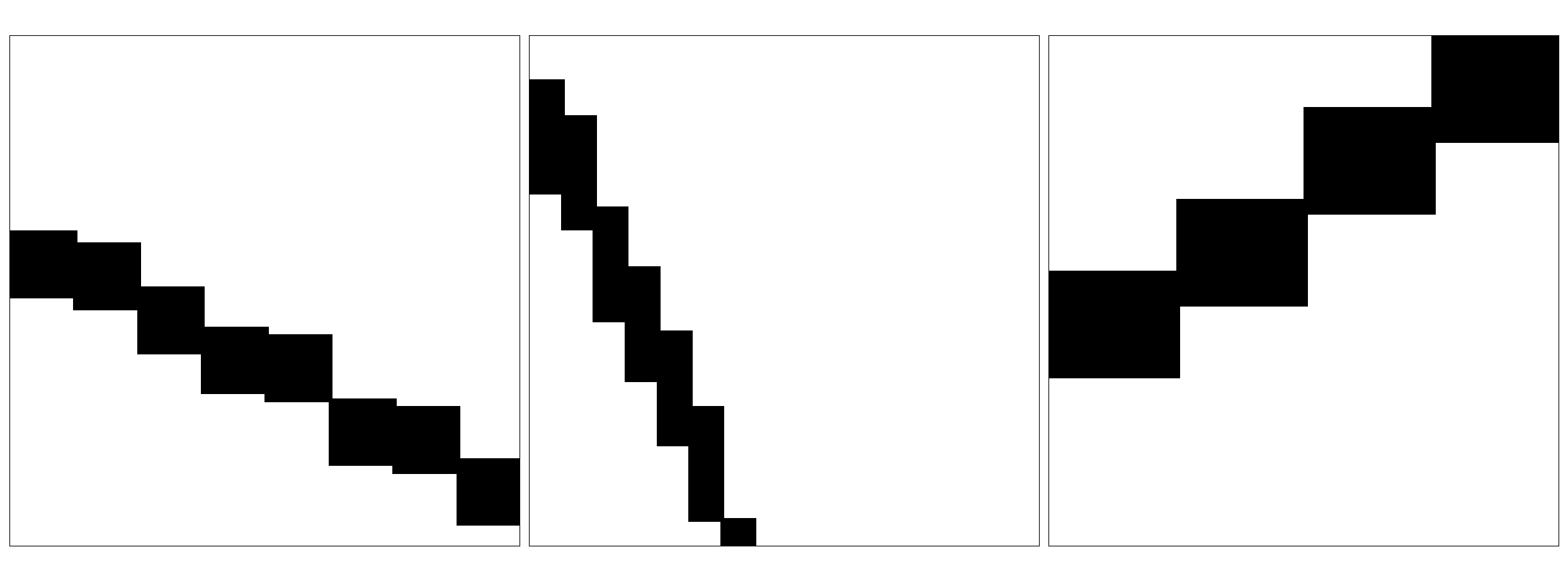}
    \caption{Example mask shapes that were used for evaluation. The masks are artificially generated to simulate common patterns that occur in data losses.}
    \label{fig:plt_mask}
\end{figure}

We run our tests on a dataset of 1000 random crops, sampled from the HiRISE terrain survey following the same procedure described in Section \ref{sec:methods_dataset}.
To conduct the experiments, we first sampled crops containing only valid pixels (i.e., without missing values) to serve as the baseline, and then generated masks to remove selected known data.
We shaped the masks in a way that resembles aliased lines, as this is a kind of artifact pattern that tends to commonly occur in Martian terrain heightmaps (refer to Figure \ref{fig:hirise} for a visual reference). We randomly generated the masks by varying different parameters of the line, such as the orientation or the number of segments. Examples of different mask shapes are shown in Figure \ref{fig:plt_mask}.

\subsection{3D Visualization}
\label{sec:experiments_3d}

\begin{figure}[H]
    \centering
    \includegraphics[width=0.75\linewidth]{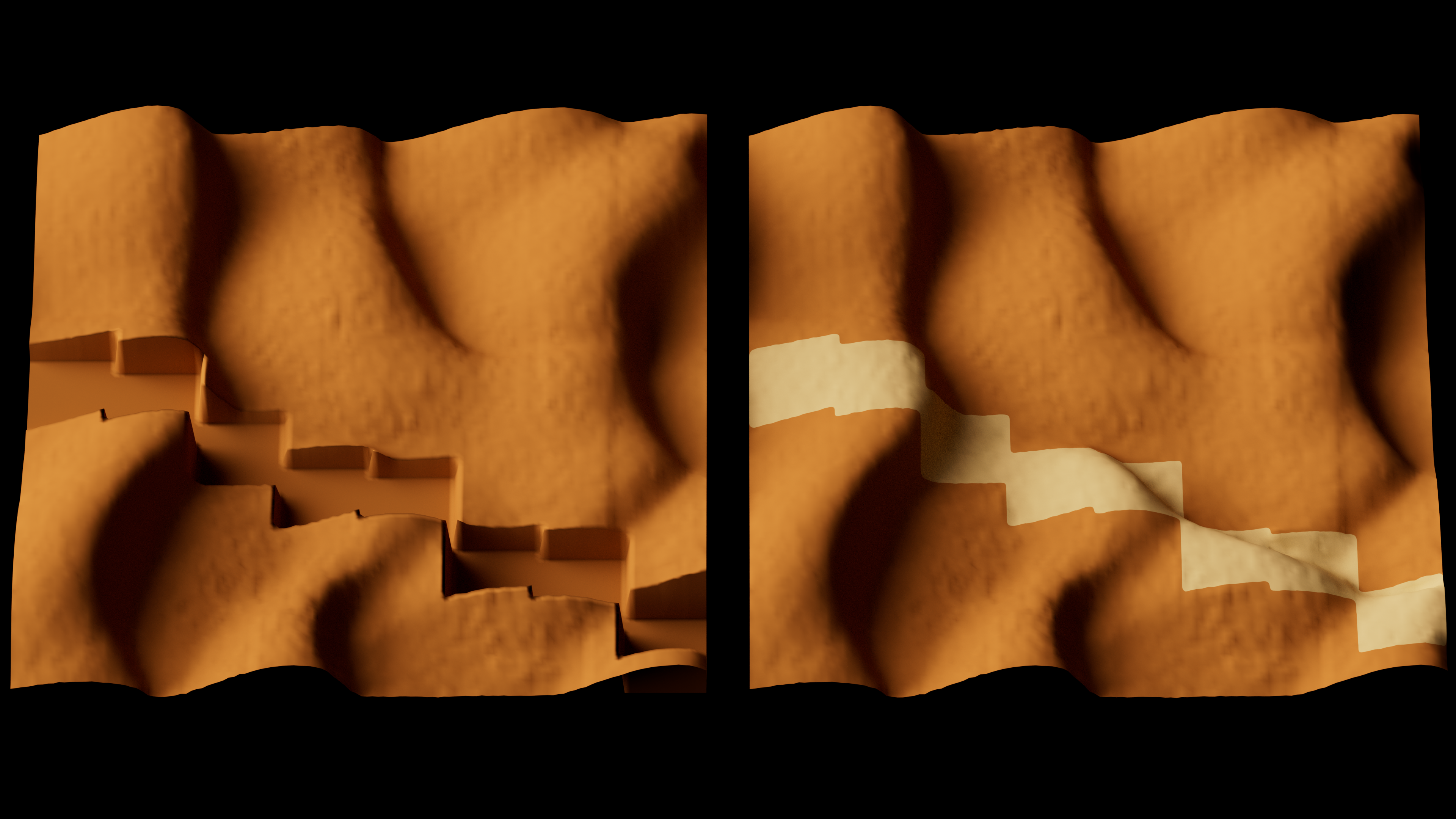}
    \caption{Input (left) and output (right) of our approach to Martian surface reconstruction, represented as 3D surfaces and rendered with Arnold in Maya.}
    \label{fig:maya_in_out}
\end{figure}

One of the main goals of this study, aimed at the investigation of \textbf{H1}, was to evaluate the restoration quality of our proposed diffusion-based method, when rendered as 3D surfaces within virtual environments. Indeed, we aimed to perform void-filling that would result in geometrically and visually coherent meshes, making the restored surfaces suitable for use in VR applications. In order to carry out an in-depth evaluation, we used Autodesk Maya\footnote{\url{https://www.autodesk.com/products/maya/overview}}, a software for 3D modeling and rendering. Its built-in Python API allowed us to automate the import process of heightmaps and create meshes for them. Additionally, we leveraged the Arnold renderer\footnote{\url{https://www.autodesk.com/products/arnold/overview}} to create custom shaders capable of highlighting restored areas based on the binary masks, and to produce detailed visualizations under varying lighting conditions for qualitative analysis (Figure \ref{fig:maya_in_out}).

\subsection{Comparative Evaluation}
\label{sec:experiments_comparative}

In order to assess the performance of our method for Mars surface restoration (\textbf{H2}), other inpainting and void-filling algorithms were evaluated for comparison. The scenario of terrain restoration led us to choose Inverse Distance Weighting (IDW) \cite{Shepard1968} and kriging \cite{McBratney1986} as our primary baselines, given their widespread adoption in this context. We also included an image inpainting technique in our tests, namely the Navier-Stokes algorithm \cite{Bertalmio2001}. All methods were tested on the same dataset and with the same set of masks. The implementation for kriging was provided by the PyKrige library\footnote{\url{https://geostat-framework.readthedocs.io/projects/pykrige/en/stable/index.html}}, whereas OpenCV\footnote{\url{https://opencv.org/}} was used for Navier-Stokes. For IDW we used $N = 12$ neighbours and a power parameter of $p = 2$.


\begin{figure*}[t!]
    \centering
    \includegraphics[width=\textwidth]{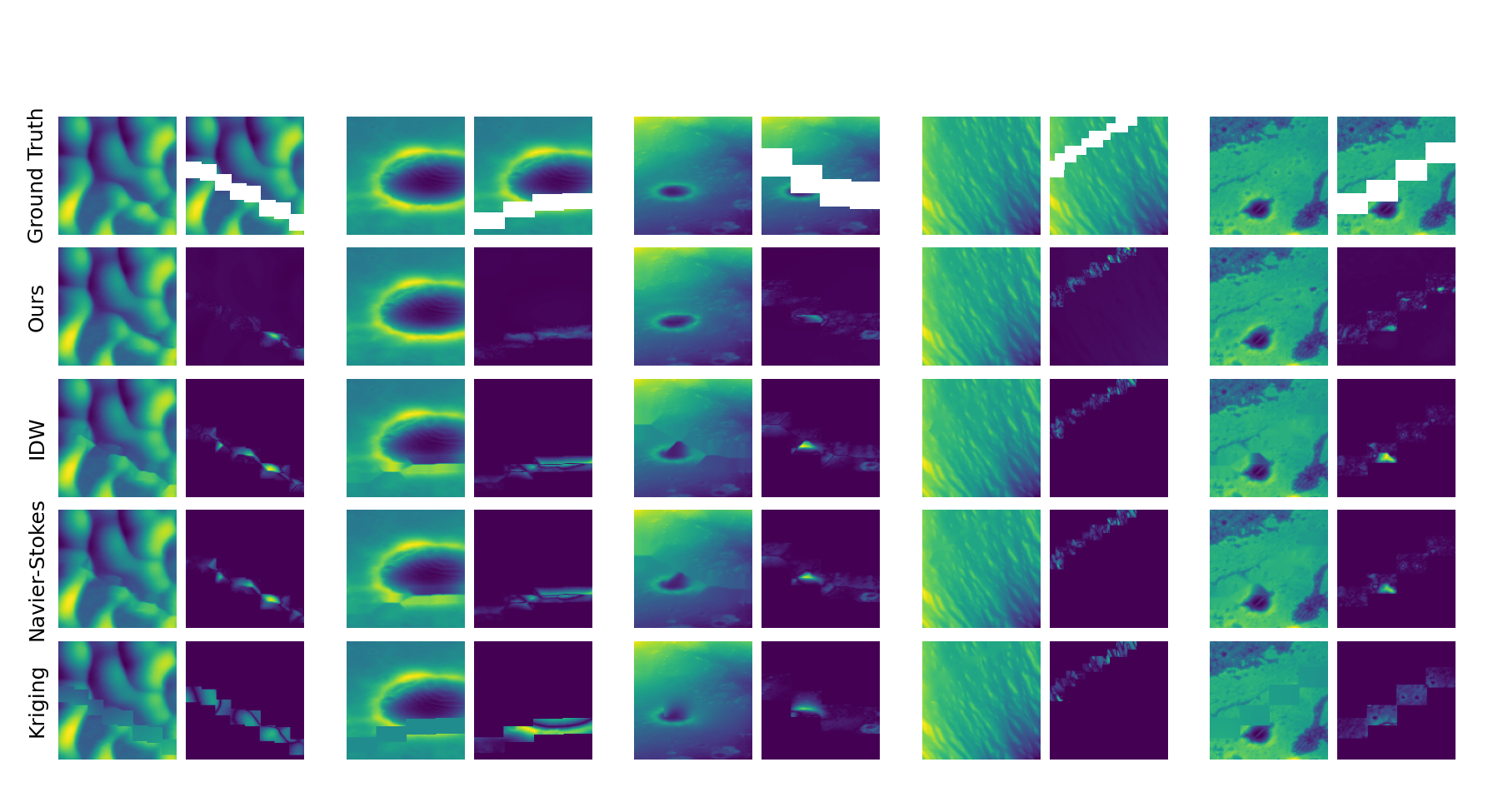}
    \caption{Qualitative evaluation of results, represented as 2D rasters. For each group, the right column represents the absolute error between restored and original values; the lack of yellow-green pixels represents better reconstructions.}
    \label{fig:qualitative_plt}
\end{figure*}

\begin{figure*}[t!]
    \centering
    \includegraphics[width=\textwidth]{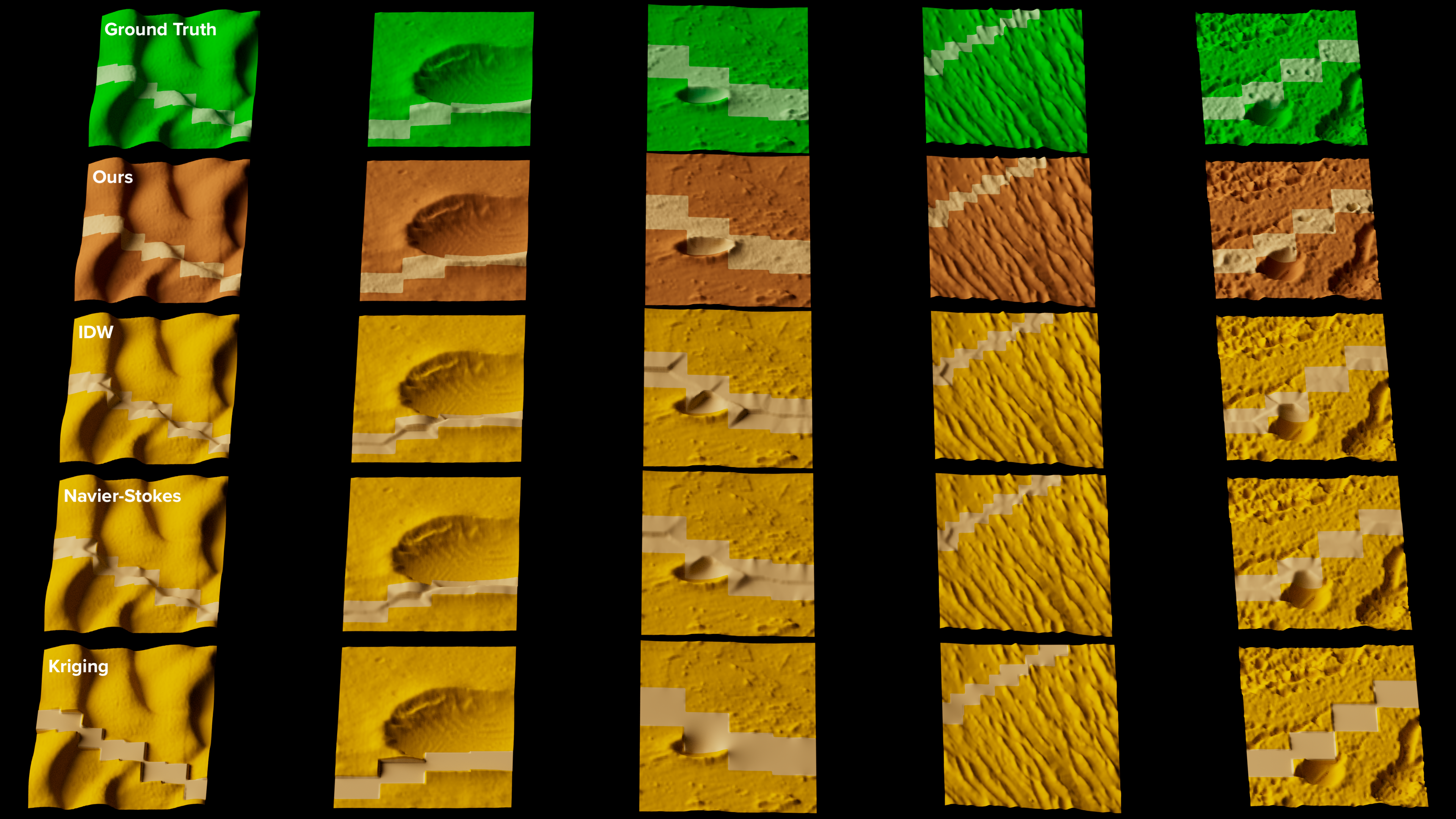}
    \caption{Qualitative evaluation of results, represented as 3D surfaces and rendered with Arnold in Autodesk Maya. The highlighted parts indicate the areas reconstructed using the evaluated methods. The possibility to control lighting and visualization conditions allows to perform effective assessments of the results.}
    \label{fig:qualitative_maya}
\end{figure*}


\section{Results}
\label{sec:results}

\subsection{Visual Assessment}
\label{sec:results_visual}

The visual inspection of the tested surface reconstruction methods was conducted using both 2D plots and 3D renders. They highlighted significant structural differences between diffusion-based restoration and interpolation algorithms, with the former producing more coherent results, in support of \textbf{H1}. Figure \ref{fig:qualitative_plt} displays a selection of results as color-mapped 2D images; in each example, the right column shows the absolute error with respect to the ground truth. 
These considerations become even more evident in Figure \ref{fig:qualitative_maya}, which shows the same selection of results, processed as heightmaps and rendered as 3D surfaces instead.

\subsection{Error Metrics Evaluation}
\label{sec:results_error}

We evaluated the tested methods from a quantitative perspective as well, in order to assess the reconstruction quality in terms of error with respect to the ground truth. This analysis aimed to verify whether the visually convincing results correspond to surfaces that are indeed similar to the original ones. Such validation is of critical importance: these terrains, which are frequently used for scientific analyses and mission planning purposes, must approximate their real-world counterparts as closely as possible. In this context, for each tested sample we computed the Root Mean Squared Error (RMSE), Mean Absolute Error (MAE), and Peak Signal-to-Noise Ratio (PSNR) between the ground truth and the restoration output. In addition, we also considered Earth Mover's Distance (EMD) as a measure of the discrepancy between value distributions.

Table \ref{tab:error} reports the average results over the 1000 random surface crops that were tested. Our method outperforms the others across all metrics, suggesting that the effectiveness observed during visual assessments is also measured quantitatively, supporting \textbf{H2}. Kriging, which leverages information related to spatial correlation among valid pixels, is the second-best performing, yielding results that are close to those of our approach. It also interesting to note that the Navier-Stokes algorithm, typically used for image inpainting but less common in the surface void-filling domain, performs generally better than a well-known method in this context such as IDW.

\subsection{Perceptual Metrics Evaluation}
\label{sec:results_perceptual}

The quantitative evaluation was further extended to include the calculation and comparison of perceptual metrics, a family of measures that are designed to model the similarity between images, as perceived by humans. Evaluations of this kind are commonly performed in image generation tasks, as they aim to approximate how closely the generated outputs resemble the training distribution, attempting to model human perceptual judgments in assessing whether the generated images ``look like'' the original ones. The selected metrics are the following:
i) Structural Similarity Index Measure (SSIM) \cite{ZhouWang2004}, which evaluates image similarity in terms of luminance, contrast, and structural consistency;
ii) Learned Perceptual Image Patch Similarity (LPIPS) \cite{Zhang2018}, which computes the distance between latent feature representations of two images as extracted by a pre-trained neural network;
iii) Fr\'echet Inception Distance (FID) \cite{Heusel2017}, which tests which compares the distribution of synthetic images against that of real images using activations from an Inception network. Although the results may be biased by the network being trained on the ImageNet dataset, we decided to employ it in addition to the others as a further evaluation method.
It is worth pointing out that, while some perceptual metrics are inherently related to error-based ones, they are often sensitive to different types of visual discrepancies \cite{Hore2010}.

Table \ref{tab:perceptual} presents the average results for SSIM and LPIPS across the tested 1000 samples; the FID score is a single value for each method, obtained by comparing the overall distribution of the results against the distribution of ground truth images. Our method achieves superior results in both LPIPS and FID, whereas kriging yields a slightly higher average SSIM score; on this front, \textbf{H2} was supported for the most part, while also giving further insights in support of \textbf{H1}. Notably, Navier-Stokes continues to outperform IDW in this evaluation as well.

\section{Discussion}
\label{sec:discussion}

The methodology we presented showed to be an effective approach for reconstructing missing sections of Martian terrains.

\textbf{Visual assessment.} 
Visual assessments of the reconstructed heightmaps, conducted both using 2D color-mapped rasters and 3D renderings, provided valuable insights. We paid particular attention to fidelity and structural consistency, especially near the boundaries between valid and missing regions. Our method proved effective with these regards, as it blends the two areas seamlessly and generates realistic surface patches with smooth transitions, while at the same time reconstructing patterns that were already present. IDW and Navier-Stokes exhibit similar behaviour between them, as both expand the values of valid pixels into the missing regions. Navier-Stokes tends to appear slightly smoother than IDW, likely due to the fluid dynamics principles on which the algorithm is based. Both methods show discontinuities near the center of the reconstructed regions, which is particularly suboptimal when considering these terrains as testbeds for rover simulations or morphological studies. Kriging also yields interesting results: it performs effectively on flat regions and adapts with reasonable accuracy to some complex structures, but it struggles to capture high-frequency patterns such as dunes or crater ridges; however, the resulting artifacts generally have a less disruptive impact on the overall surface structure than those produced by IDW and Navier-Stokes. These findings support hypothesis \textbf{H1}, confirming that the restored surfaces exhibit high structural fidelity and accuracy.


\begin{table*}[t!]
\centering

\begin{tabular}{lcccc}
\hline
& \textbf{RMSE} $\downarrow $ & \textbf{MAE} $\downarrow$ & \textbf{PSNR} $\uparrow$ & \textbf{EMD} $\downarrow$ \\
\hline
\textit{Ours} & \textbf{0.0752} (--) & \textbf{0.0548} (--) & \textbf{39.4494} (--) & \textbf{0.1366} (--) \\
IDW & 0.0864 (+14.9\%) & 0.0603 (+10\%) & 35.7504 (-9.4\%) & 0.1756 (+28.6\%) \\
Navier-Stokes & 0.0830 (+10.4\%) & 0.0580 (+5.8\%) & 36.2110 (-8.2\%) & 0.1699 (+24.4\%) \\
Kriging & 0.0784 (+4.3\%) & 0.0573 (+4.6\%) & 39.3492 (-0.3\%) & 0.1572 (+15.1\%) \\
\hline
\end{tabular}

\caption{Average results of the error-based metrics for the test run (1000 samples). Bold values indicate the best results.}
\label{tab:error}
\end{table*}

\begin{table*}[t!]
\centering

\begin{tabular}{lccc}
\hline
& \textbf{SSIM} $\uparrow$ & \textbf{LPIPS} $\downarrow$ & \textbf{FID} $\downarrow$ \\
\hline
\textit{Ours} & 0.9660 (--) & \textbf{0.0754} (--) & \textbf{10.9397} (--) \\
IDW & 0.9620 (-0.4\%) & 0.1365 (+81\%) & 61.0935 (+458.5\%) \\
Navier-Stokes & 0.9650 (-0.1\%) & 0.1181 (+56.6\%) & 42.0554 (+284.4\%) \\
Kriging & \textbf{0.9684} (+0.2\%) & 0.0973 (+29\%) & 27.5676 (+152\%) \\
\hline
\end{tabular}

\caption{Average results of the perceptual-based metrics for the test run (1000 samples). FID is not an average, but a single score associated to each evaluation set. Bold values indicate the best results.}
\label{tab:perceptual}
\end{table*}


\textbf{Objective Measurements.}
Our method produces reconstructions that are also accurate in terms of objective accuracy with regards to the original terrains. It achieved the best performance across multiple error metrics, namely Root Mean Squared Error (RMSE), Mean Absolute Error (MAE), Peak Signal-to-Noise Ratio (PSNR) and Earth Mover's Distance (EMD). This quantitative analysis verified hypothesis \textbf{H2}, which advocated for a reliable objective restoration of missing Mars regions. 
Finally, we evaluated a series of perceptual similarity metrics that are commonly employed to model human perception in the context of generative models. Our method obtained the best scores in Learned Perceptual Image Patch Similarity (LPIPS) and Fréchet Inception Distance (FID), whereas kriging achieved the highest Structural Similarity Index Measure (SSIM) score. This outcome might stem from the structure comparison component that is present in SSIM computations, which may penalize reconstructions that are structurally coherent yet mismatched (as produced by our method) compared to flatter, less detailed restorations (as generated by kriging). These results point to a solid and robust framework, and to an additional overall confirmation of \textbf{H2}, despite not having reached the best performance on all fronts. The relatively small size of the training dataset may be a factor with this regard, however the domain of extraterrestrial terrains is inherently data-scarce and even in these conditions we achieved notable results.

\textbf{Qualitative vs. Quantitative Insights.}
The relationship between qualitative evaluation and quantitative metrics offers further valuable insights. Despite the relatively strong performance of kriging, which achieved the best scores among interpolation-based methods, and also outperformed our method in terms of SSIM, the visual assessment clearly showed that the diffusion-based approach provides with more realistic and coherent reconstructions. This can be largely attributed to the statistical nature of kriging, a model which computes an autocorrelation semivariogram to specifically minimize the mean squared error of the approximation function, and further underlines the importance of subjective evaluation of the results.
Our initial hypotheses have been thoroughly addressed and verified. We provided insights into the capabilities of diffusion models to encode large amounts of data, and applied them to reconstruct heightmaps in ways that are well-suited for being represented as accurate 3D virtual terrains. Results indicated that our trained model represents a highly flexible tool for surface reconstruction at different scales. Moreover, we showed that unconditional models in particular can be successfully leveraged for restoration purposes, without having to rely on additional information aside from the raster. These results open up promising scenarios for the application of deep generative models in extraterrestrial contexts, characterized by similarly limited data availability.

\textbf{Limitations.}
The current work presents some limitations. 
First, both the model architecture and the overall inpainting pipeline offer substantial room for optimization. Users do not have the possibility to define custom crop regions or binary masks in real time, that limits the ability to perform precise and localized inpainting during immersive sessions. Currently, this remains a technical barrier due to the computational cost of the method.
Comparisons with other diffusion-based approaches \cite{Lo2024} still need to be evaluated. These methods have been mostly developed for geoscientific applications and tested on Earth-related tasks; despite our emphasis on the unconditional nature and the the generalization potential of our approach, they make for valuable testing grounds nonetheless. 
The applicability of this method to other extraterrestrial bodies, such as the Moon, has yet to be assessed. The general framework of augmenting existing terrain surveys can be adapted to diverse scenarios, though each must be customized to the specific environmental and data constraints. 

\textbf{Future Work.}
There are several directions in which the current work can be further expanded. As a next step, we plan to conduct a user study to evaluate the perceived quality and usability of the reconstructed terrains from the perspective of participants. This may include assessments of the Sense of Presence, interaction techniques with the reconstructed surfaces in scientific space exploration simulations, and navigation methods within the virtual Martian environments.
Regarding our inpainting method, multi-resolution models and upgraded architectures such as Denoising Diffusion Implicit Models (DDIMs) \cite{Song2020}, Latent Diffusion Models (LDMs) \cite{Rombach2022} or Diffusion Transformers \cite{Peebles2023} may enable faster and more precise generations, along with the possibility to employ this method in diverse practical uses. Moreover, alternative configurations of the RePaint algorithm can be explored in order to achieve more efficient inpainting results, ideally with a smaller number of required steps. Such functionalities would require a significantly faster inference process, allowing user input to guide the reconstruction dynamically, potentially within immersive environments. 
In addition, this line of research could contribute to the development of experimental setups for the human evaluation of AI-generated content in VR environments. While current studies in this area primarily focus on 2D imagery \cite{Yang2024}, we believe the current work has potential for expanding these evaluations to 3D contexts.

\textbf{Ethics and Replicability}
Ethics Committee approval was not required for this study, as the experiments did not involve human participants or any living beings. Furthermore, the inference was performed on publicly available open data, ensuring transparency and replicability. The processed data are available upon request.

\section{Conclusions}
\label{sec:conclusions}

This work presented a method for the restoration of degraded Martian heightmaps using unconditional diffusion models. We trained a DDPM on a dataset of terrains sampled at various scales, enabling the model to learn common morphological patterns across multiple resolutions. The approach was compared with established void-filling and inpainting algorithms, with the aim of evaluating both the structural fidelity of the reconstructions for immersive 3D visualization in VR, and the accuracy of the restored Martian surfaces in the absence of additional conditioning information. Visual inspection showed that the generated regions preserve structural consistency, while quantitative metrics confirmed a close match with the original data in terms of reconstruction error and perceptual similarity. Notably, the results indicate strong potential for integrating these reconstructions into VR applications for space simulation and exploration, supporting our initial hypotheses. 

\bibliographystyle{unsrt}
\bibliography{main}

\end{multicols}

\end{document}